\documentclass[10pt,journal,compsoc]{IEEEtran}
\newif\ifpeerreview

\peerreviewtrue

\usepackage[nocompress]{cite}
\usepackage{url}
\usepackage{amsmath,amssymb,graphicx}
\usepackage{lipsum} 
\usepackage[switch]{lineno}
\usepackage{mathtools}
\usepackage{commath}
\usepackage[ruled,vlined]{algorithm2e}
\usepackage{setspace}
\usepackage{natbib}
\usepackage{authblk}
\usepackage{hyperref}

\newcommand\given[1][]{\:#1\vert\:}
\usepackage{textcomp}
\def\Var{{\textrm{Var}}\,}
\newcommand\tab[1][1cm]{\hspace*{#1}}

\title{SimuGAN: Unsupervised forward modeling and optimal design of a LIDAR Camera}

\author[1]{Nir Diamant}
\affil[1]{Computer Science Department, Technion - IIT, Israel}
\author[2]{Tal Mund}
\author[2]{Ohad Menashe}
\author[2]{Aviad Zabatani}
\author[1]{Alex M. Bronstein}
\affil[2]{RealSense, Intel, Israel}

\begin{document}

\IEEEtitleabstractindextext{%
\begin{abstract}
  Energy-saving LIDAR camera for short distances estimates an object's distance using temporally intensity-coded laser light pulses and calculates the maximum correlation with the back-scattered pulse.
  Though on low power, the backs-scattered pulse is noisy and unstable, which leads to inaccurate and unreliable depth estimation.
  To address this problem, we use GANs (Generative Adversarial Networks), which are two neural networks that can learn complicated class distributions through an adversarial process. We learn the LIDAR camera's hidden properties and behavior, creating a novel, fully unsupervised forward model that simulates the camera. Then, we use the model's differentiability to explore the camera parameter space and optimize those parameters in terms of depth, accuracy, and stability. To achieve this goal, we also propose a new custom loss function designated to the back-scattered code distribution's weaknesses and its circular behavior. The results are demonstrated on both synthetic and real data.   
  
\end{abstract}

}

\ifpeerreview
{}
\fi
\maketitle


\IEEEraisesectionheading{
  \section{Introduction}}\label{sec:introduction}
  3D depth-sensing devices have become ubiquitous in the last decades, finding applications ranging from robotics, 3D scanning, autonomous driving, drones, and agriculture, biology, geology, atmosphere, to military uses, law enforcement, and even gaming. The LIDAR technology has a wide market on its own, and its active laser illumination makes it resilient to low-light environmenta and harsh weather conditions such as rain and fog. A LIDAR camera illuminates the scene using a laser and senses the scattered light using a photo-detector. The LIDAR camera's performance and accuracy depend on external parameters such as the distance and reflectivity of the target and various internal parameters, which control its inner components' behavior. For example, the laser driver electronics affect the laser's raising time, falling time, and the general shape of the transmitted signal. Another example is the working point of the photo-detector, which controls the amplification of the received signal.\\
  
  To calculate an object's distance, a binary code pulse, denoted $\alpha$, which is a sequence of light pulses, is transmitted by the LIDAR camera transmitter. The pulse hits the target and reflects to the receptor (we denote the received signal by $\alpha'$). The most accurate time of light travel back and forth, denoted $t_{argmax}$, is the peak of the correlation between $\alpha$ and $\alpha'$. Denoting the receiver's sampling rate by $f$, the speed of light by $c$, and the system delay\footnote{The system delay might change as a function of the camera parameters, and used in the function as an averaged anchor} in mm by $d$, the estimated distance of the object is given by $\Delta =\dfrac{1}{2} \cdot (\dfrac{t_{argmax}}{f} \cdot c-d)$, ignoring triangulation corrections.

However, the method described above is hard to rely on: The back-scattered pulse of bits behaves differently than the transmitted one. The behavior is affected by the distance, the camera's parameters, ambient light, thermal noise, and even the previously transmitted bits.
this unexplained behavior causes uncontrollable inaccuracies of the depth estimation.   
\begin{figure}[h]
\centering
\includegraphics[width =\columnwidth]{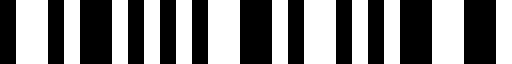}
\caption{The real transmitted code: Each horizontal row is a set of pixels representing transmitted code which was transmitted 64 times in a row.
}
\end{figure}
\begin{figure}[h]
\centering
\includegraphics[width =\columnwidth]{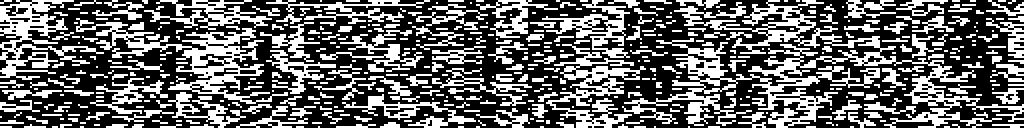}
\caption{Demonstration of the instability of the back-scattered code: Each horizontal row is a set of pixels representing the back-scattered code sampled 64 times with the same
transmitted code and camera parameters. It can be seen that the code is phase shifted from the original transmitted code. finding a reliable estimation of the  distance from the object is hard due to the amount of noise and inconsistency between the different rows in the back-scattered codes, which makes the correlation between the transmitted code and the back-scattered one not reliable}
\end{figure}
  
  Since the number of parameters is vast, finding the optimal set of parameters for a specific task might be challenging. In many cases, each camera component is optimized separately to achieve maximal performance in terms of SNR. However, once combined, the optimal parameters per component might not be optimal in terms of the camera's performance. The optimal working point of the depth camera depends on the application it is used for, as well as the scene itself and the objects we wish to capture. 3D scanning, for example, will require maximal depth accuracy, while autonomous driving applications might settle on lower accuracy given object detections in longer distances.  
  For a given objective, searching the space of all possible parameter values for its optimum is costly, inefficient, and impractical when the space is sufficiently large. To find such an optimum, one might utilize standard mathematical optimization techniques such as gradient approaches. However, an analytical formulation of the gradient in respect of the parameters might not be available, and numerical gradients calculations become costly as the number of parameters grows. Besides, some parameters might only obtain discrete values, which cause inaccurate numerical gradient estimations. \\
  To push research on laser-based semantic segmentation, \cite{behley2019semantickitti} annotated all KITTI Vision Odometry Benchmark, with a provided 360\textdegree{} field of view of dense LIDAR depth points, to generate three new benchmark related tasks. The use of such an extensive LIDAR depth annotated data can serve many purposes. i.e., \cite{wu2017squeezeseg} leveraged it for teaching a deep network model to do real-time road object semantic segmentation. \\ 
 
  Generative Adversarial Networks (GANs) are two different neural networks, Generator and Discriminator, that train against each other in an adversarial process.
  The Discriminator is being fed with real independent data and with the generator model's output, and its goal is to distinguish between real and fake data.
  The Generator, on the other hand, tries to fool the Discriminator and make it classify its outputs as real ones.
  The end goal of this process is two well-trained models: The Discriminator is an expert in deciding which samples are drawn from the real data distribution, while the Generator is an expert in generating new unreal samples that have the same distribution of the real data, as described in  \cite{gans,dcgans}. 
  An essential capability of the GANs is to create an output conditioned by some attributes fed as an additional input to both models.
  The Generator will then generate different classes of the original distribution, depending on the condition. Such model is called "CGAN". \cite{cgans}.\\
  
  We will show hereby how to use this ability of GANs for learning how LIDAR back-scattered code (the signal that returns from the object to the camera detector) behaves and distributes as a function of the camera parameters. Our model learns an interpolation and extrapolation of the parameters space to receive a continuous and differentiable forward function out of discrete real-world parameters options. We then optimize the parameters to receive an optimal calibration of the camera.
  We focus on short-range distance objects and low power consuming LIDAR for energy-saving and eye-care products.

%
%
%
%



\section{Related Work}

Improving LIDAR performance is crucial for enhancing the depth measurements for many of the mentioned uses.
\cite{xu2019depth} used a model that receives a LIDAR scan with an RGB image for input by modeling the geometric constraints between the surface normal and depth and combining this information with a calculation of the confidence level of sparse LIDAR measurements to lower the impact of noise.
\citet*{8959801} optimized the six external calibration parameters with an algorithm that aims to align the edges detected in the LIDAR point cloud with the corresponding color image, combined with minimizing the depth difference between the measured LIDAR data and a depth map derived from a monocular image. 
Another enhancement to the LIDAR performance was done by \cite{9105252}, involving deep learning methods to optimize the depth estimation of dense 3D point clouds, which on its own is limited when sampling with low rates.
Different uses of signal generation using GANs were previously done: \cite{8852227} and \cite{hartmann2018eeggan} developed a generative model to generate EEG-like brain signals for purposes like data augmentation, supersampling, and restoration of corrupted data; all of those can be similarly used in the field of sensing and computational imaging. 
\cite{caccia2019deep} touched both aspects in their research and generated a 2D point map out of LIDAR scans, using GANs, reaching high-quality samples.
To improve depth estimation sparsity of pixels, acquired by a LIDAR, \cite{yang2019dense} inferred the posterior distribution of an image depth map, to have a probability over depth for each pixel in the image.\\
The depth estimation field contains different methods to do so and represents 3D image reconstruction out of a specific scene. \cite{DBLP:journals/corr/LainaRBTN16} constructed a fully convolutional network with residual blocks that models the ambiguous mapping between monocular images and depth maps and helps estimate the depth map of a scene given only a single RGB image. \cite{DBLP:journals/corr/abs-1804-06278} took it one step further and, with a deep network, predicted the reconstruction of a piece-wise planar depth map. The model receives a single RGB image for input and infers the set of plane parameters and corresponding plane segmentation masks.
\cite{Aleotti2018GenerativeAN} used GANS for monocular depth estimation, with the following method: the Generator infers a depth map given an input of a 2D image, while the Discriminator has to distinguish between the Generator's output and target frames acquired with a stereo rig.

\section{Proposed Method}
Our research splits into two different parts: 
\begin{enumerate}
\item Creating a fully unsupervised Forward model of the LIDAR camera, that learns the behavior and distribution of the back-scattered signals and their reliance on the camera's parameters.\\
\item Defining a robust metric and creating a fully unsupervised Inverse model allows us to traverse smoothly through the parameters space and find the optimal camera calibration.
\end{enumerate}

\subsection{Part 1: Creating a fully unsupervised forward model using CGAN}
To understand the behavior of the camera's light radar (LIDAR), we collected an extensive database of back-scattered codes related to a specific object, and the same code transmitted, but with a variety of different parameters of the camera, having multiple samples per a set of parameters to capture the distribution of the different behaviors.
\subsubsection{Training a Semi-Supervised CGAN}
we trained a conditional GAN to distinguish between real back-scattered codes and fake ones, conditioned on a set of different eight continuous parameters of the camera normalized to be in the range of [0,1], denoted $C$.
The Generator model receives an input a set of parameters of the camera, and some random noise $z\sim \mathcal{N}(0,1)$ for keeping the output stochastic. The Generator outputs an alleged code that is plausible to be back-scattered from the object to the camera's receptor.\\
Neither the ground truth distance of the object nor the real transmitted code is being fed to the network and learned inherently during the training phase, making the learning fully unsupervised.
\\
The Discriminator model, unlike the original CGAN model, receives only a single input vector, without any condition, and outputs two different objects, similar to what was done in \cite{8803807}:
\begin{enumerate}
\item Validity score: How confident is the Discriminator whether the code vector is a real back-scattered code from the object.
\item Parameters estimation: An eight continues parameters vector, which is the most plausible to cause such a back-scattered code, denoted $\hat{C}$.
\end{enumerate}

\begin{figure*}[h]
\centering
\includegraphics[width = 1\textwidth]{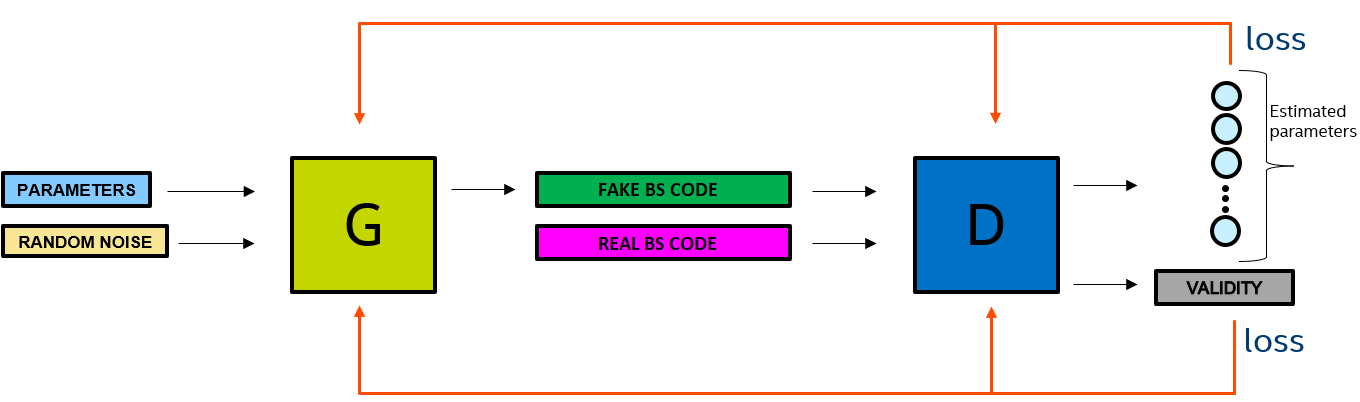}
\caption{Diagram of our CGAN model - Forward model of the camera. 
"BS" stands for "Back-Scattered"}
\end{figure*}

\subsubsection{Custom losses}
For stabilizing the adversarial process, we used a Wasserstein GAN \cite{wgans} and applied another loss for the Discriminator, namely "Gradient Penalty" loss, which enforces the Discriminator Gradient's second norm to be bounded by 1. \cite{DBLP:journals/corr/GulrajaniAADC17}
Additionally, for the Discriminator, we added an MSE loss over the predicted parameters, $C'$, encouraging them to be similar to those set while $\alpha$ was transmitted. \\
For the Generator, we added three new terms, additionally to the basic adversarial loss, divided into two types:
\begin{enumerate}
\item An adversarial loss for the predicted camera parameters that the Discriminator outputs, $\hat{C}$, when having the code that the Generator generates as input, compared to the camera parameters that were fed as input to the Generator, $C$.
\item A loss of a first norm over the distribution of the $1_{st}$ and $2_{nd}$ moments of the distribution of the bits generated by the Generator, comparing to the $1_{st}$ and $2_{nd}$ moments of the distribution of the bits in the real data, respectively, having the same camera parameters, $C$. \\ \\
To further stabilize the gentle joint distribution of the conditions and the generated data of the CGAN, we first trained the Generator, having only a loss consisting of the $1_{st}$ and $2_{nd}$ moments of the distributions differences, and after several epochs, we gradually increased the weight of the adversarial losses. The notion derives from curriculum learning, where the neural network is being fed first on easy samples, and step by step learns to handle greater and more complicated tasks, as shown in \cite{10.1145/1553374.1553380} \\
\end{enumerate}
$D_{loss} =\underbrace{ {\mathbb E}[D(G(z\given C))[0]]}_{\text{Adversarial loss}} - \underbrace{{\mathbb E}[D(x\given C)[0]]}_{\text{loss over real samples}}\\ \tab + \underbrace{\lambda_{GP} \cdot {\mathbb E} [(\norm{  \nabla_{x}D(x\given C)[0]} _{2}-1)^2]}_{\text{Gradient penalty loss}}\\ \tab +\underbrace{\lambda_{parameters} \cdot \norm{D(x\given C)[1]-C}_{2}}_{\text{Predicted conditions loss}}
$
\\ \\ \\

$G_{loss} = \underbrace{-{\alpha \cdot \mathbb E}[D(G(z\given C))[0]]}_{\text{Adversarial loss}} \\ \tab + \underbrace{\alpha \cdot \lambda_{parameters} \cdot \norm{D(G(z\given C)))[1]-C}_{2}}_{\text{Adversarial predicted conditions loss}} \\ \tab+ \underbrace{\lambda_{mean} \cdot \norm{{\mathbb E}[G(z \given C)] - {\mathbb E}[x \given C]}}_{\text{1}_{\text{st}}\text{ Moment distribution similarity loss}} \\ \tab + \underbrace{\lambda_{variance} \cdot \norm{\Var{}[G(z \given C)] - \Var{}[x \given C]}}_{\text{2}_{\text{nd}}\text{ Moment distribution similarity loss}}$
\\ \\ \\
where:\\ $D(x)=\text{[-validity score, predicted parameters]}$\\
$\alpha= 
\begin{cases*}
    \text{min}\left(1, \dfrac{\text{\#iterations}}{\text{Constant}}\right) & \text{\#iterations}$> \#threshold$ \\
    0 & otherwise
    \end{cases*}$\\

\subsubsection{Visualization of the training phase}
To visually track the convergence process, we plotted along with the training, the real back-scattered codes, and the generated ones by our CGAN of the five most stable and unstable camera parameters of the training data. Each row represents such binary code, and the codes that correspond to the same camera parameters are stacked together to notice how similar they are. A large difference between two consecutive rows indicates a high variance and noisy back-scattered codes. The cleaner and more constant rows we get will indicate a more stable back-scattered code reflected in a better and more reliable estimation of the depth. Using this visualization, we could track both the correctness of the output distribution, alongside the behavior of the back-scattered codes conditioned by the camera's parameters.\\

\begin{figure}[h]
\includegraphics[width = \columnwidth]{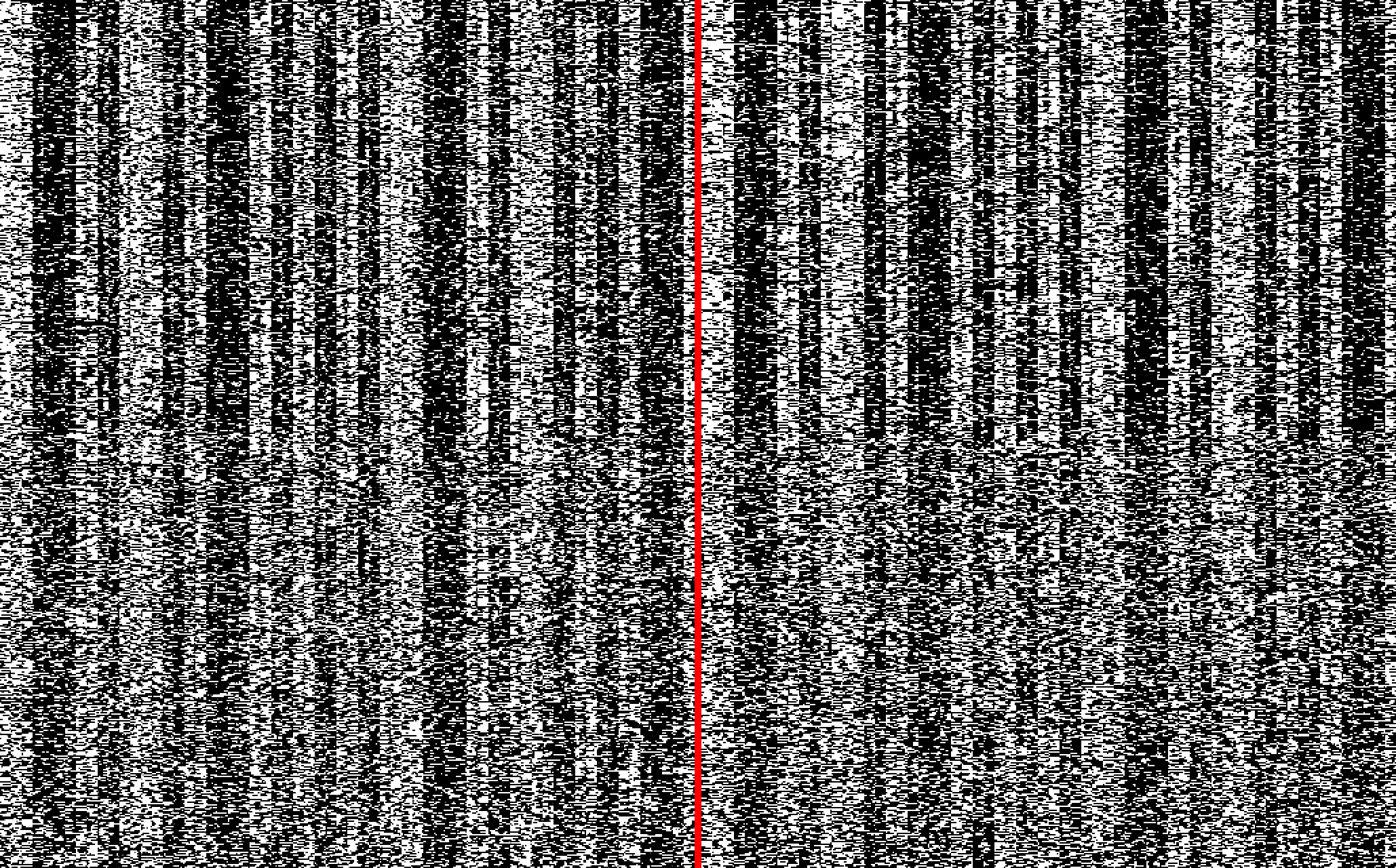}
\caption{The outputs of the Generator are on the left side of the real back-scattered codes, corresponding to the same camera parameters. The upper half of each side is the generated/back-scattered codes of the five most stable parameters, reflecting more concrete and clear columns, which means the code is stable. the lower halves are the generated/back-scattered codes of the five most unstable parameters, reflecting by a much noisier output with a higher variance and unclear data.}
\end{figure}

\subsubsection{Models architecture}
Both Generator and Discriminator consist of 10 1-dimensional convolution blocks with only two channels and 7-sized kernels for all layers, which means that the receptive field captured 67 bits around the area of each bit transmitted. That allows the model to consider the causal connections between different hidden effects that happen along a sequence of a transmitted pulse of code. The padding method is circular since the transmitted and back-scattered codes are cyclic. The nonlinear function at the end of each block is LeakyRelu, with a slope of 0.2 at nonpositive values.
The Generator's last block ends with a Sigmoid activation to push all values to be 0 or 1 as they are supposed to represent a back-scattered binary code.
The Generator's first layer and the Discriminator's last layer are fully connected layers for adjusting the correct output size needed.

\section{Part 2: Creating a fully unsupervised inverse model}
The trained CGAN, which imitates the LIDAR outputs distribution, let us have a differentiable forward model.
Nevertheless, we need to have a reliable metric to decide what an optimal set of parameters is. It is enough to have one extreme noisy sample, namely "outlier," in a batch of back-scattered codes to dramatically affect the mean and variance of the estimated depth. That makes the metric of checking the mean and variance of the same bits of back-scattered codes corresponding to the same transmitted code and camera parameters not sufficient and even unreliable in terms of successful depth estimation. 
To overcome this problem, We formed a new metric to evaluate the back-scattered code quality,
"Inliers rate", denoted '$R$' to be the following: \\   $R = 100 \cdot \dfrac{\#\text{samples s.t.} \norm{\Delta - median(\Delta)} \leq delta_{in}  }{\#\text{samples}}$, where $delta_{in}$ was set to be 0.03m.\\

\subsection{The optimization algorithm}
The camera parameters are designed to be ranged in $[0,1]$ after normalization, yet the Generator model can receive an unbounded input; hence, the optimization process's output might exceed the boundaries. \\
To solve this issue, we set the optimization object to be the inverse function of a Sigmoid $\left(f(x) = \dfrac{1}{1+ e^{-\text{x}}}\right)$ of the parameters. Every iteration of the optimization process, the optimization object passes through a Sigmoid to receive the current step camera parameters in the range of [0,1].\\
We then calculate the correlation $\rho$ and find its peak using a differentiable approximation of the $argmax$ function to receive the time travel $t_{argmax}$ of light to and back from the object. The approximation assumes a normal distribution of the outputs of the correlation. It is implemented by the sum of a softmax function of the values, multiplied by the set of integer numbers ranged from 1 to the code length $L$. \\
For the optimization loss metric, we chose a combination of two different losses:
\begin{enumerate}
    \item $loss_{median}$: The median of all the estimated distances along the batch of optimization shall be a reliable candidate when the data is not too noisy. Hence we treated the median as our anchor. since there are a few extreme outliers, we used the natural log of the difference between the distance and the median distance as our loss. This encourages all the results to crowd together around the median without suffering from heavy penalties caused by a few samples.
    to adopt this to the LIDAR cyclic code transmission, we took the distance to be the minimum of the former term and the difference between of the former term to the maximal distance that the camera is capable of calculating. \\
    \item $loss_{variance}$: To accurate the results, we added another loss that encourages the variance of the same bits along the batch axis to be as low as possible, what promises a more stable result.
\end{enumerate}
The optimization algorithm used SGD with a stop condition of $R > 97\%$. 
Assuming the batch output distributes normally, $\Delta \sim \mathcal{N}(median(\Delta),\sigma^2)$, the objective of the optimization function will get to optimum as $\sigma\to 0$. \\
Along all the optimization procedure, the object's distance was not introduced to the system, and the model gets to its goal of minimizing the distribution variance of the estimated distances around the real distance in an unsupervised manner. 
\begin{algorithm}[!h]
\setstretch{1.5}
\caption{finding the optimal camera's parameters}
$\text {Denoting the Batch Size} = n$: \\


$\text{opt} \xleftarrow{} x\sim \mathcal{N}(0,1)^{\text{num parameters}}$ \\
$\widetilde{\text{opt}} \xleftarrow{} \ln{\frac{\text{opt}}{1-\text{opt}}}$\\
\While{$R \leq th$}{
$\text{opt} = \dfrac{1}{1+ e^{-\text{opt}}} $ \\
$\rho = corr(\alpha,\text{ }  G(z|\text{opt}))$ \\
$t_{argmax} = \sum_{i=1}^{L} \dfrac {e^{\rho_{j}}}{\sum_{j=1}^{L} e^{\rho_{j}}}\cdot[L]$ \\
$\Delta =\dfrac{1}{2} \cdot \left(\dfrac{t_{argmax}}{f} \cdot c-d\right)$ \\
 $loss_{median} = \dfrac{1}{n} \cdot \sum_{i=1}^{n} \log(1+\min(\norm{\Delta-median(\Delta)},\dfrac{1}{2} \cdot \left(\dfrac{L}{f} \cdot c-d\right)- \norm{\Delta-median(\Delta)}))$ \\
$loss_{variance} = var(G(z|\text{opt}))$ \\
$loss = \beta \cdot loss_{median} + (1- \beta) \cdot loss_{variance}$ \\
$\widetilde{\text{opt}} = \widetilde{\text{opt}} - \lambda_{LR} \cdot  \dfrac{\partial {loss}}{\partial {\widetilde{\text{opt}}}}$ \\
$R = \% \norm{\Delta - median(\Delta)} \leq \delta_{in}$
} return opt
\end{algorithm}

\section{Experimental Results}
Running the whole algorithm pipeline on Both simulated data and real camera data has shown a significant improvement in the distribution of the estimated distances of the objects.
Training the CGAN over the real data was a much challenging task, due to the many real-world factors that intervene in the system and dramatically affect the joint distribution of the camera parameters and the object's scenario and characteristics.\\
All experiments were conducted on hard and noisy examples, i.e., low laser power, which makes the algorithm useful for efficient LIDAR uses, consuming low power, saving on energy, and significantly lowering the danger of harming the human eye. \\
The training time of the CGAN takes only several hours on a single GPU, and the optimization process takes several seconds to half a minute to find the optimal camera parameters.\\

The following diagrams show the difference of the distance estimation distribution before and after using our algorithm, transmitting the same code 50,000 times with one constant set of random camera parameters and then with the optimized parameters. The depth estimation distribution with a random set of camera parameters is broad, with many outliers. After applying our algorithm, the optimal parameters perform a much stable depth estimation, reflecting in a much narrower distribution around the real object's distance, with almost no outliers.  \\

\begin{figure}[h]
\centering
\includegraphics[width = \columnwidth]{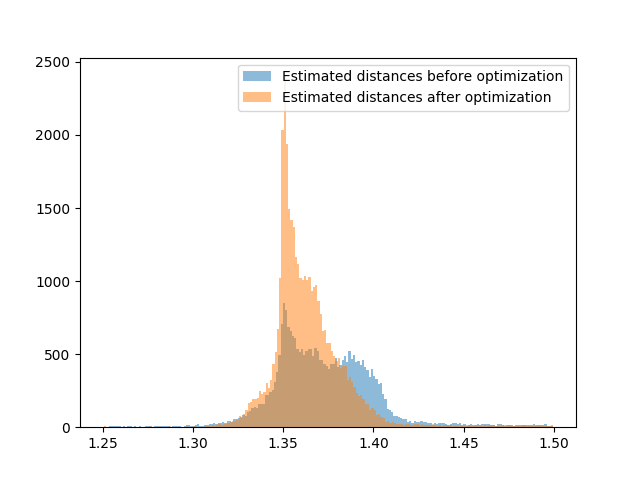}

\caption{distribution of estimated distances, before and after using our algorithm on simulated data.}
\end{figure}

\begin{figure}[h]
\centering
\includegraphics[width = \columnwidth]{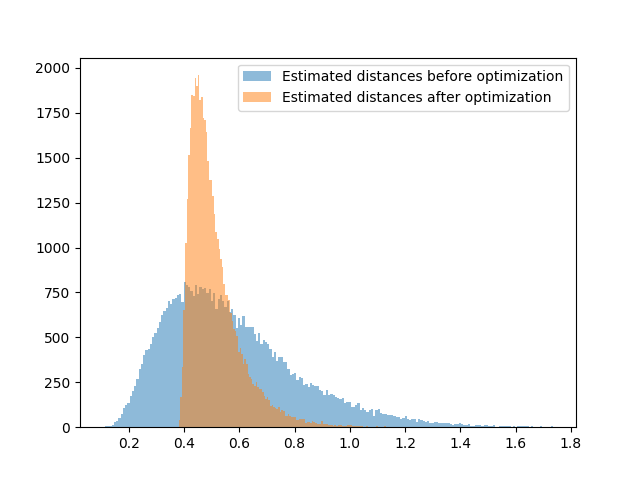}
\caption{distribution of estimated distances, before and after using our algorithm on real data}
\end{figure}



The following image shows two batches of 512 back-scattered codes:  on the left, at the beginning of the optimization procedure, and on the right, at the end. On each side, all the rows correspond to the same set of parameters, which differ between the sides. Both sides correspond to the same transmitted binary code. It can be seen that before the optimization phase, the variance between the rows is high and that the density of each white/black part is lower, which implies incorrect and distorted back-scattered code.\\ After the optimization procedure, the rows are much similar to each other, having a much solid group of bits. The back-scattered code is much more similar to the transmitted one, and hence the depth estimation distribution is significantly narrower and more reliable.

\begin{figure}[h]
\includegraphics[width = \columnwidth]{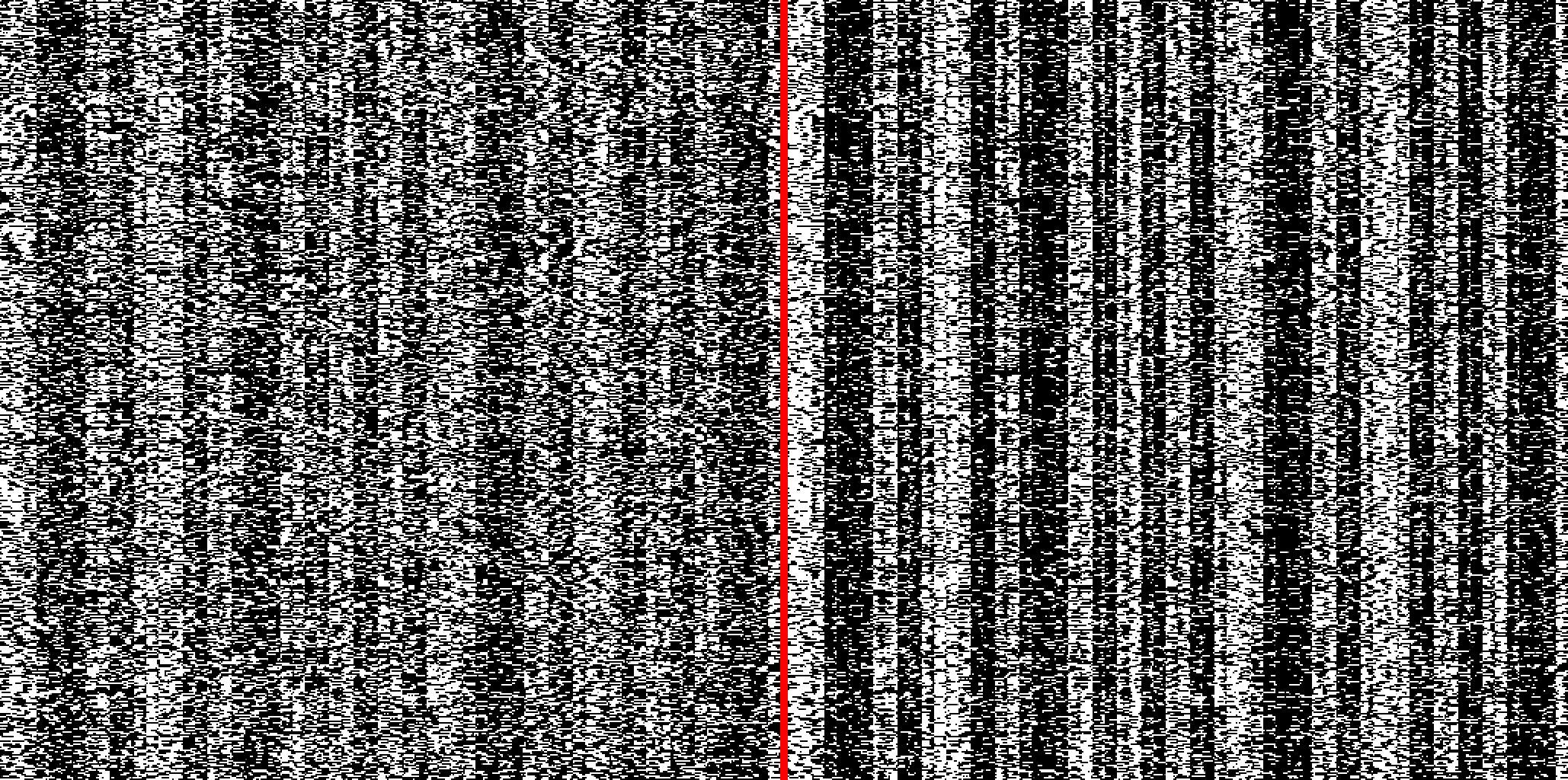}

\caption{Comparing the back-scattered code quality - before and after the optimization process}
\end{figure}
\section{Conclusion}
A framework for optimizing the working point of a LIDAR camera in order to enhance its depth estimation stability was introduced. The framework consists of two optimization processes: (i) Teaching the Generator the forward model and thereby creating a simulator of the real system. (ii) Optimizing the parameters of the camera to minimize the noise in the depth estimation. The resulting Generator can be a valuable tool as a simulator with the important qualities of fast inference and derivable output in respect to input. 
The framework is flexible. For any given starting point and an objective, we can optimize the parameters and achieve better performance. The concept itself can be expanded to various systems in which a group of parameters controls the system's behavior. For the specific use case of depth estimation stability, having a derivable forward model opens new possibilities when dealing with learning tasks. Given a neural network that receives the back-scattered signal as input and reconstructs the depth, one could jointly learn the reconstruction network and the camera parameters.\\  
We have shown that a Generator can serve as a fast simulator for a real system, aided by the $1_{st}$ and $2_{nd}$ moments losses to help capture the low order statistics and the Discriminator for the higher orders. Besides, we aided by the Wasserstein GAN technique and condition reconstruction to avoid mode collapse and learn the joint distribution. \\
Also, the Generator can easily optimize the system's parameters, given a derivable objective, Improving current conditions(camera parameters) to a better set of conditions. Once given a trained Generator, which can be treated as a flexible block, replacing the objective may be easy. \\
Looking at the big picture, the same concept is not limited to 3D imaging devices. It can be expanded for other systems with a large number of parameters and a derivable objective. Besides, this GAN training can create a useful simulator for other tasks (that do not require optimization).\\
An open question is how much data each system requires for training. We shall use a smaller dataset and still get a qualitative simulator of real product behavior. \\
Optional future work may do On-The-Fly optimization of camera parameters given distances and reflectivities in scenes. Another enhancement would be replacing the depth estimation with a Neural Network model and optimizing the model with the conditions simultaneously.

\ifpeerreview \else
\section*{Acknowledgments}
The authors would like to thank...
\fi

\bibliography{references}

\ifpeerreview \else


\begin{IEEEbiography}{Michael Shell}
Biography text here.
\end{IEEEbiography}


\begin{IEEEbiographynophoto}{John Doe}
Biography text here.
\end{IEEEbiographynophoto}


\fi

\end{document}